\documentclass[]{fairmeta}

\usepackage{amsmath}
\usepackage{amsfonts}
\usepackage{hyperref}
\usepackage{url}
\usepackage{booktabs}
\usepackage{multirow}
\usepackage{pgfplots}
\usepackage{graphicx}
\usepackage{array}
\pgfplotsset{compat=1.18}
\usepackage{arydshln}

\makeatletter
\def\adl@drawiv#1#2#3{%
        \hskip.5\tabcolsep
        \xleaders#3{#2.5\@tempdimb #1{1}#2.5\@tempdimb}%
                #2\z@ plus1fil minus1fil\relax
        \hskip.5\tabcolsep}
\newcommand{\cdashlinelr}[1]{%
  \noalign{\vskip\aboverulesep
           \global\let\@dashdrawstore\adl@draw
           \global\let\adl@draw\adl@drawiv}
  \cdashline{#1}
  \noalign{\global\let\adl@draw\@dashdrawstore
           \vskip\belowrulesep}}
\makeatother

\newcommand{\pk}{\texttt{Memory}}
\newcommand{\pkplus}{\texttt{Memory+}}

\title{Memory Layers at Scale}
\author[*]{Vincent-Pierre Berges}
\author[*]{Barlas O\u{g}uz}
\author[]{Daniel Haziza}
\author[]{Wen-tau Yih}
\author[]{Luke Zettlemoyer}
\author[]{Gargi Ghosh}

\affiliation[]{Meta FAIR}

\contribution[*]{Main authors}

\abstract{Memory layers use a trainable key-value lookup mechanism to add extra parameters to a model without increasing FLOPs.  Conceptually, sparsely activated memory layers complement compute-heavy dense feed-forward layers, providing dedicated capacity to store and retrieve information cheaply.  This work takes memory layers beyond proof-of-concept, proving their utility at contemporary scale.  On downstream tasks, language models augmented with our improved memory layer outperform dense models with more than twice the computation budget, as well as mixture-of-expert models when matched for both compute and parameters.  We find gains are especially pronounced for factual tasks.  We provide a fully parallelizable memory layer implementation, demonstrating scaling laws with up to 128B memory parameters, pretrained to 1 trillion tokens, comparing to base models with up to 8B parameters.}

\date{\today}
\correspondence{Vincent-Pierre Berges at \email{vincentpierre@meta.com}, Barlas O\u{g}uz at \email{barlaso@meta.com}}

\metadata[Code]{\url{https://github.com/facebookresearch/memory}}
\metadata[Blogpost]{\url{https://ai.meta.com/blog/meta-fair-updates-agents-robustness-safety-architecture}}

\begin{document}

\maketitle
\section{Introduction}
\label{section:intro}
\begin{figure}[h!]
    \centering
    \begin{minipage}{0.49\textwidth}
        \centering
        \begin{tikzpicture}
            \begin{axis}[
                width=7cm, height=6cm,
                xlabel={Memory Parameters ($\times 2048$)},
                ylabel={Accuracy},
                xmode=log,
                log basis x={10},
                xtick={262144,  1048576, 4194304, 16777216, 67108864},
                xticklabels={$0$, $1024^2$, $2048^2$, $4096^2$, $8192^2$},
                ymin=0, ymax=70,
                xmin=200000, xmax=90000000,
                grid=major,
                legend pos=north west,
                title={Factual QA Accuracy vs. Memory Size}
            ]

            \addplot[color=blue, mark=square*, line width=1pt] 
            coordinates {
                (262144, 7.728531856) (1048576, 9.833795014)
                (4194304, 14.43213296) (16777216, 20.13850416)
                (67108864, 20.77562327)
            };
            \addlegendentry{NQ}
            
            \addplot[color=red, mark=triangle*, line width=1pt] 
            coordinates {
                (262144, 32.72696511 ) (1048576,39.47388513)
                (4194304, 51.1213206 ) (16777216, 58.98492884 )
                (67108864, 62.1361763 )
            };
            \addlegendentry{TQA}

            \addplot[dashed, color=blue, line width=1pt] 
            coordinates {(10, 25.097) (90000000, 25.097)};

            \addplot[dashed, color=red, line width=1pt] 
            coordinates {(10, 64.003) (90000000, 64.003)};

            \end{axis}
        \end{tikzpicture}
    \end{minipage}
    \hfill
    \begin{minipage}{0.49\textwidth}
        \centering
        \begin{tikzpicture}
            \begin{axis}[
                width=7cm, height=6cm,
                xlabel={Memory Parameters ($\times 2048$)},
                ylabel={Negative Log Likelihood},
                xmode=log,
                log basis x={10},
                xtick={262144, 1048576, 4194304, 16777216, 67108864},
                xticklabels={$0$,  $1024^2$, $2048^2$, $4096^2$, $8192^2$},
                ymin=3, ymax=13,
                xmin=200000, xmax=90000000,
                grid=major,
                legend pos=north east,
                title={NLL vs. Memory Size}
            ]
            
            \addplot[color=blue, mark=square*, line width=1pt] 
            coordinates {
                 (262144, 11.75866084 ) (1048576, 10.52263569 )
                (4194304, 9.35401662 ) (16777216, 8.549570421 ) (67108864, 8.035854722 )
            };
            \addlegendentry{NQ}
            
            \addplot[color=red, mark=triangle*, line width=1pt] 
            coordinates {
                (262144, 6.830461913 ) (1048576, 6.255764939 )
                (4194304, 5.07787363 ) (16777216,4.414831458) (67108864,4.121618934)
            };
            \addlegendentry{TQA}

            \addplot[dashed, color=blue, line width=1pt] 
            coordinates {(10, 7.767) (67108864, 7.767)};

            \addplot[dashed, color=red, line width=1pt] 
            coordinates {(10, 3.966) (67108864, 3.966)};
            

            \end{axis}
        \end{tikzpicture}
    \end{minipage}
    \caption{Scaling the size of the memory for a 1.3 billion parameter base model (zero memory parameters corresponds to a dense model), trained to 1 trillion tokens.  On the left, factual QA accuracy (exact match on NaturalQuestions and F1 score on TriviaQA), on the right task NLL (lower is better). Dashed lines show the performance of a 7B model trained on 2 trillion tokens with 10x more FLOPs.}
    \label{fig:memory_scaling}
\end{figure}
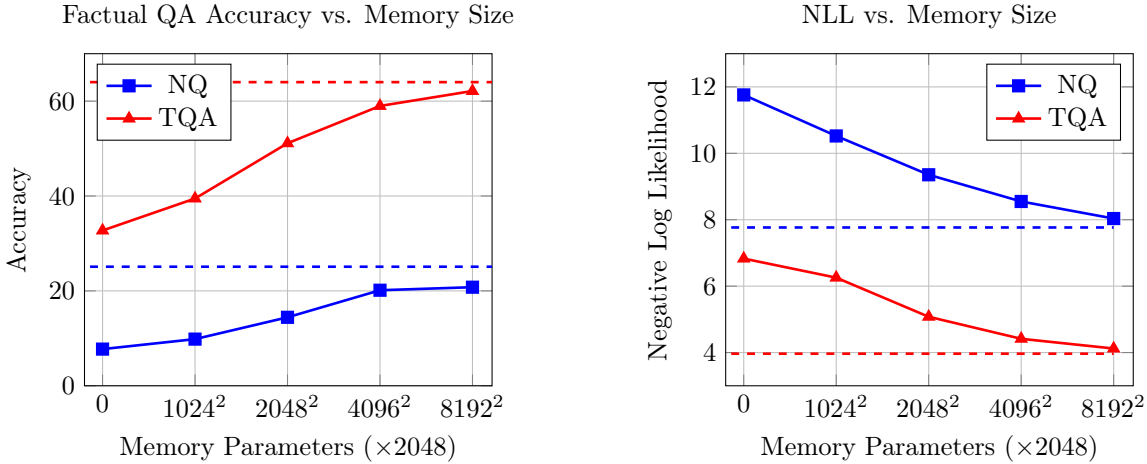
Pretrained language models encode vast amounts of information in their parameters~\citep{roberts2020knowledgepackparameterslanguage}, and they can recall and use this information more accurately with increasing scale~\citep{brown2020languagemodelsfewshotlearners}.  For dense deep neural networks, which encode information primarily as weights of linear matrix transforms, this scaling of parameter size is directly coupled to an increase in computational and energy requirements.  It is unclear if this is the most efficient solution to all information storage needs of language models.  An important subset of information that language models need to learn are simple associations.  For example, LLMs learn birthdays of celebrities, capital cities of countries, or how one concept might relate to another.  While feed-forward networks can in principle (given sufficient scale) learn any function~\citep{HORNIK1989359}, including lookup tables of associations, using an associative memory for this purpose would be both more efficient and more natural.

Such memory layers can be implemented with a simple and cheap key-value lookup mechanism where both keys and values are encoded as embeddings~\citep{weston2015memorynetworks}.  Earlier works introduced end-to-end trainable memory layers~\citep{sukhbaatar2015endtoendmemorynetworks} and incorporated them as part of neural computational systems~\citep{graves2014neuralturingmachines}.  Despite early enthusiasm however, memory layers have not been studied and scaled sufficiently to be useful in modern AI architectures.  There are distinctive challenges one encounters when attempting to scale memory layers, which we touch upon in~\cref{architectures}.  In contrast to dense layers which are predominantly FLOP-bound, memory layers with their sparse activation pattern are almost entirely memory bandwidth bound.  Such components are rarely used in modern architectures and have not been optimised for hardware accelerators.  In addition to, and partly as a result of this, little research was done to improve their performance.  Instead, the field focused on alternatives such as mixture-of-experts~\citep{shazeer2017outrageouslylargeneuralnetworks}, which more closely resemble dense networks and are thus easier to scale.

In this work, we show that memory layers, when improved and scaled sufficiently, can be used to augment dense neural networks to great benefit. We do so by replacing the feed-forward network (FFN) of one or more transformer layers with memory layers (we leave other layers unchanged).  These benefits are consistent across a range of base model sizes (ranging from 134 million to 8 billion parameters), and memory capacities (up to 128 billion parameters).  This represents a two orders of magnitude leap in memory capacity compared to previous memory layers reported in the literature.  Our results (\cref{experiments}) indicate that memory layers improve the factual accuracy of language models by over 100\% as measured by factual QA benchmarks, while also improving significantly on coding (HumanEval, MBPP) and general knowledge (Hellaswag, MMLU).  In many cases, memory augmented models can match the performance of dense models that have been trained on 4x more compute.  They also outperform mixture-of-experts architectures with matching compute and parameter size, especially on factual tasks.  Given these findings, we strongly advocate that memory layers should be integrated into all next generation AI architectures.

\section{Related work}\label{related}
Language model scaling laws~\citep{kaplan2020scalinglawsneurallanguage} study the empirical performance of language models as they are scaled in compute, data, and parameter size.  Scaling laws are typically formulated in terms of training/test log likelihood, which is generally believed to correlate well with downstream performance.  Scaling plots on downstream tasks are also not without precedent~\citep{brown2020languagemodelsfewshotlearners}, but have sometimes been shown to exhibit non-linear behaviour and phase transitions~\citep{wei2022emergentabilitieslargelanguage,Ganguli_2022}.  Nevertheless, given a well behaved metric (such as task likelihood loss), most tasks exhibit smooth improvements with scaling~\citep{schaeffer2023emergentabilitieslargelanguage}.

\cite{kaplan2020scalinglawsneurallanguage} showed that performance scales log-linearly with compute and parameter size across a wide range of architecture hyper-parameters, such as model depth and width.  It has been difficult to find architectures which substantially deviate from these laws.  Mixture-of-experts (MOE)~\citep{shazeer2017outrageouslylargeneuralnetworks,lepikhin2020gshardscalinggiantmodels} is a notable exception.  MOE adds extra parameters to the model without increasing the computation budget.  While scaling laws for MOE also mostly focus on training perplexity, gains transfer well to downstream applications, as evidenced by the popularity of MOE architectures in recent state-of-the-art model families~\citep{jiang2024mixtralexperts,openai2024gpt4technicalreport, geminiteam2024gemini15unlockingmultimodal}.  Nevertheless, scaling laws for specific task families and capabilities like factuality remain understudied.

Like MOE, memory augmented models also aim to augment the parameter space of the model without adding significant computational cost.  Memory networks were proposed initially in~\citep{weston2015memorynetworks}, and with end-to-end training in~\citep{sukhbaatar2015endtoendmemorynetworks}.  Neural Turing Machines~\citep{graves2014neuralturingmachines, graves2016hybrid} combine external trainable memory with other components to build a neural trainable computer.  Product-key networks~\citep{lample2019largememorylayersproduct} were introduced to make the memory lookup more efficient and scalable.  The recent PEER~\citep{he2024mixturemillionexperts} builds on this work, replacing vector values with rank-one matrices, forming a bridge between memory architectures and MOE.

Factual text generation has long been considered a fundamental capability for generative models, typically benchmarked through factual open domain question answering~\citep{chen2017readingwikipediaansweropendomain, chen-yih-2020-open} and other knowledge-intensive tasks~\citep{petroni2021kiltbenchmarkknowledgeintensive}.  Being able to memorize the facts in the training corpus enables the model to answer fact-seeking, knowledge intensive tasks more factually and accurately. Indeed larger models have been shown to be more factual~\citep{roberts2020knowledgepackparameterslanguage, brown2020languagemodelsfewshotlearners}, but even modern LLMs are known to struggle with hallucination~\citep{Ji_2023}.  A tested way of ensuring more factuality is through retrieval augmented generation~\citep{lewis2021retrievalaugmentedgenerationknowledgeintensivenlp,karpukhin2020densepassageretrievalopendomain,lee2019latentretrievalweaklysupervised,guu2020realmretrievalaugmentedlanguagemodel,khandelwal2020generalizationmemorizationnearestneighbor}.  We use short-form QA tasks in this work to demonstrate the effectiveness of memory layers and leave the long-form generation tasks for future work.  Recently, a wide literature has emerged in mitigating LLM hallucinations through data related methods, architecture variants, pre-training and inference time improvements.  We refer to~\citep{Ji_2023} section 5 for a comprehensive survey.

\section{Memory Augmented Architectures}\label{architectures}
Trainable memory layers work similarly to the ubiquitous attention mechanism~\citep{bahdanau2016neuralmachinetranslationjointly}.  Given a query $q\in \mathbb{R}^n$, a set of keys $K \in \mathbb{R}^{N\times n}$ and values $V \in \mathbb{R}^{N\times n}$, the output is a soft combination of values, weighted according to the similarity between $q$ and the corresponding keys.  Two major differences separate memory layers from attention layers as they are typically used~\citep{vaswani2023attentionneed}.  First, the keys and values in memory layers are trainable parameters, as opposed to activations.  Second, memory layers typically have larger scale in terms of the number of keys and values, making sparse lookup and updates a necessity. For example in this work, we scale the number of key-value pairs to several millions.  In this case, only the top-$k$ most similar keys and corresponding values take part in the output.  A simple memory layer can be described by the following equations:
\begin{align}\label{equation1}
  I &= \text{SelectTopkIndices}(Kq),  &
  s   &= \text{Softmax}(K_{I}q),  &
  y   &= sV_I
\end{align}
 Here $I$ is a set of indices, $s \in \mathbb{R}^k$, $K_I, V_I \in \mathbb{R}^{k\times n}$, and the output $y \in \mathbb{R}^n$.  Each token embedding (for us, the output of the previous attention layer) goes through this memory lookup independently, similar to the FFN operation that we replace.

\subsection{Scaling memory layers}
Being light on compute, and heavy on memory, memory layers have distinct scaling challenges.  We detail some of these challenges and how we address them in this section.

\subsubsection{Product-key lookup}
One bottleneck which arises when scaling memory layers is the query-key retrieval mechanism.  A naive nearest-neighbour search requires comparing each query-key pair, which quickly becomes prohibitive for large memories.  While fast approximate vector similarity techniques~\citep{johnson2019billion} could be used here, it's a challenge to incorporate them when the keys are being continually trained and need to be re-indexed.  Instead, we adopt trainable product-quantized keys from~\citep{lample2019largememorylayersproduct}.  Product keys work by having two sets of keys instead of one, where $K_1, K_2 \in \mathbb{R}^{\sqrt{N} \times \frac{n}{2}}$.  The full set of keys of size $N\times n$, which is never instantiated, consists of the product of these two sets.  The top-$k$ lookup on the full set of keys can be efficiently done by searching the much smaller set of half-keys first, saving compute and memory.  To perform the lookup, we first split the query as $q_1, q_2 \in \mathbb{R}^{\frac{n}{2}}$.  Let $I_1, I_2$ and $s_1, s_2$ be the top-k indices and scores obtained from the respective key sets $K_1, K_2$.  Since there are only $\sqrt{N}$ keys in each set, this operation is efficient.  The overall indices and scores can be found by taking $\text{argmax}_{i_1\in I_1, i_2 \in I_2} s_1[i_1] + s_2[i_2]$.

\subsubsection{Parallel memory}
\begin{figure}[h!]
\includegraphics[width=\textwidth]{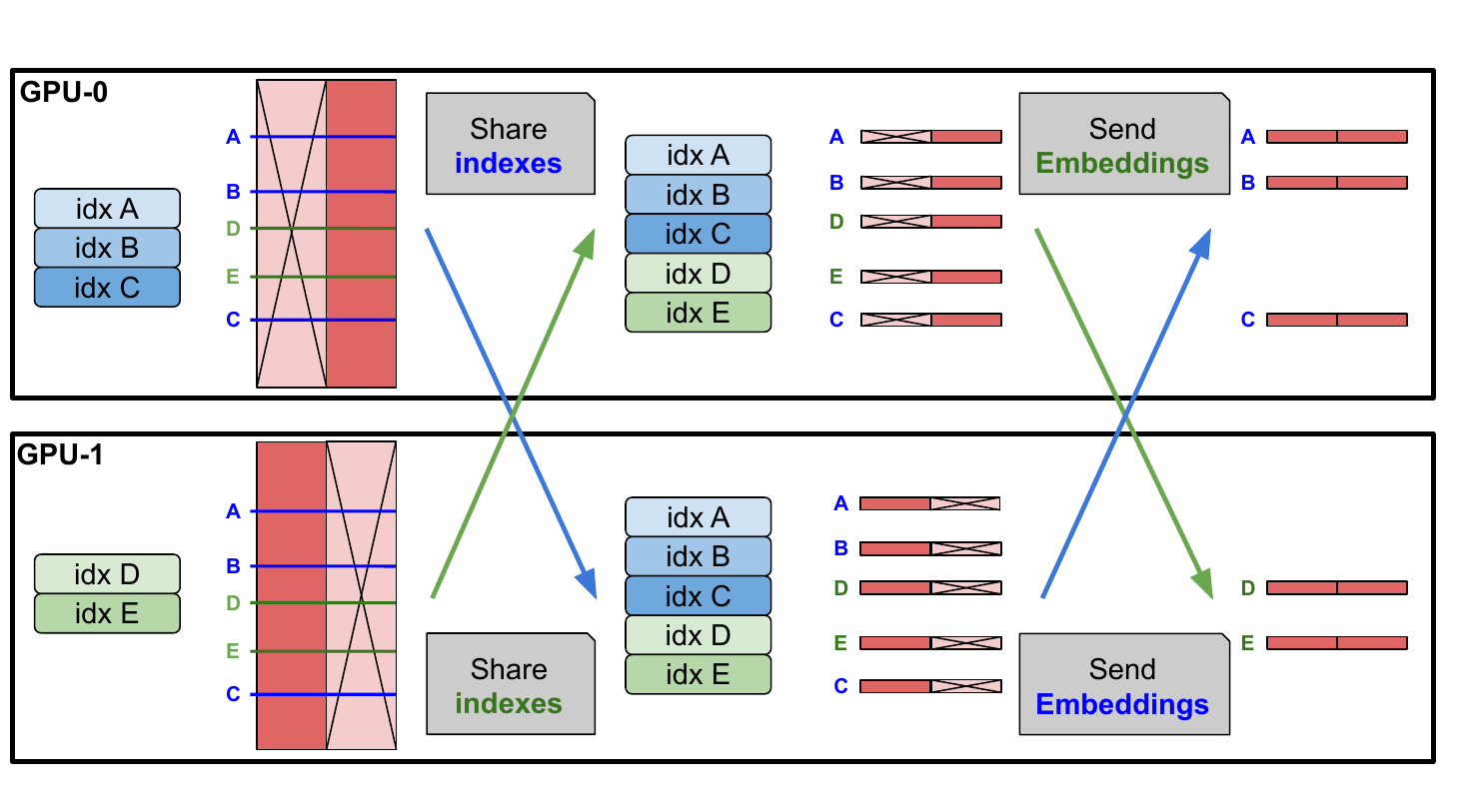}
\caption{Illustration of the parallel EmbeddingBag implementation for a ``Memory Group'' of two GPUs. Each GPU performs the EmbeddingBag operation on all of the indices of the group, but on half-dimension embeddings it has access to.}
\label{fig:memory_parallel}
\end{figure}

Memory layers are naturally memory-intensive, mostly due to the large number of trainable parameters and associated optimizer states.  To implement them at the scale of several millions of keys, we parallelize the embedding lookup and aggregation across multiple GPUs.  The memory values are sharded across the embedding dimension.  At each step, the indices are gathered from the process group, each worker does a lookup and then aggregates the portion of embeddings in its own shard.  After this, each worker gathers the partial embeddings corresponding to its own portion of the indices.  We take care to keep activation memory manageable at this stage, by making sure each GPU only gets its own portion, and does not need to instantiate the entire embedding output. The process is illustrated in~\cref{fig:memory_parallel}.  The implementation is independent of other model parallelism schemes such as tensor, context or pipeline parallelism, and operates on its own process group.  


\subsubsection{Shared memory}
Deep networks encode information at different levels of abstraction across different layers.  Adding memory to multiple layers may help the model use its memory in more versatile ways.  In contrast to previous work~\citep{lample2019largememorylayersproduct}, we use a shared pool of memory parameters across all memory layers, thus keeping parameter count the same and maximizing parameter sharing.  We find that multiple memory layers increase performance significantly over having a single layer with the same total parameter count, up to a certain number of layers (in our case, 3).  Beyond this point, replacing further FFN layers degrades performance, showing sparse and dense layers are both needed and likely complementary (see~\cref{sec:ablations}).


\subsubsection{Performance and stability improvements}
The main operations in the memory layer is to compute the weighted sum of the top-k embeddings: it is implemented in PyTorch's \texttt{EmbeddingBag} operation. As the number of floating-point operations is negligible, we expect this operation to be solely limited by the GPU memory bandwidth, but find multiple inefficiencies in PyTorch's implementation in practice. We implemented new and more efficient CUDA kernels for this operation. Our forward pass optimizes memory accesses and achieves 3TB/s of memory bandwidth, which is close to our H100 specification of 3.35TB/s (compared to less than 400GB/s with PyTorch's implementation). The backward pass is more complicated as multiple output gradients have to be propagated to the same weight gradient. We benchmarked multiple strategies: accumulation via atomic additions ("atomics"), row-level atomic lock where we amortize the cost of memory lock over the embedding dimension ("lock"), and atomic-free ("reverse\_indices"). The latter approach requires some preprocessing to inverse the token\_id to embedding\_id mapping, so that each row in the embedding gradient can know which token will contribute to it.  Typically, while the "atomics" approach is already up to 5x faster than the existing PyTorch operator, we found that the "reverse\_indices" and "lock" approaches can be faster when the embedding dimension exceeds 128, as long as the embeddings are roughly balanced.
Overall, our custom kernels make the embedding bag operation end-to-end 6x faster compared to PyTorch's \texttt{EmbeddingBag} for our use cases.

\begin{figure}[h!]
\includegraphics[width=\textwidth]{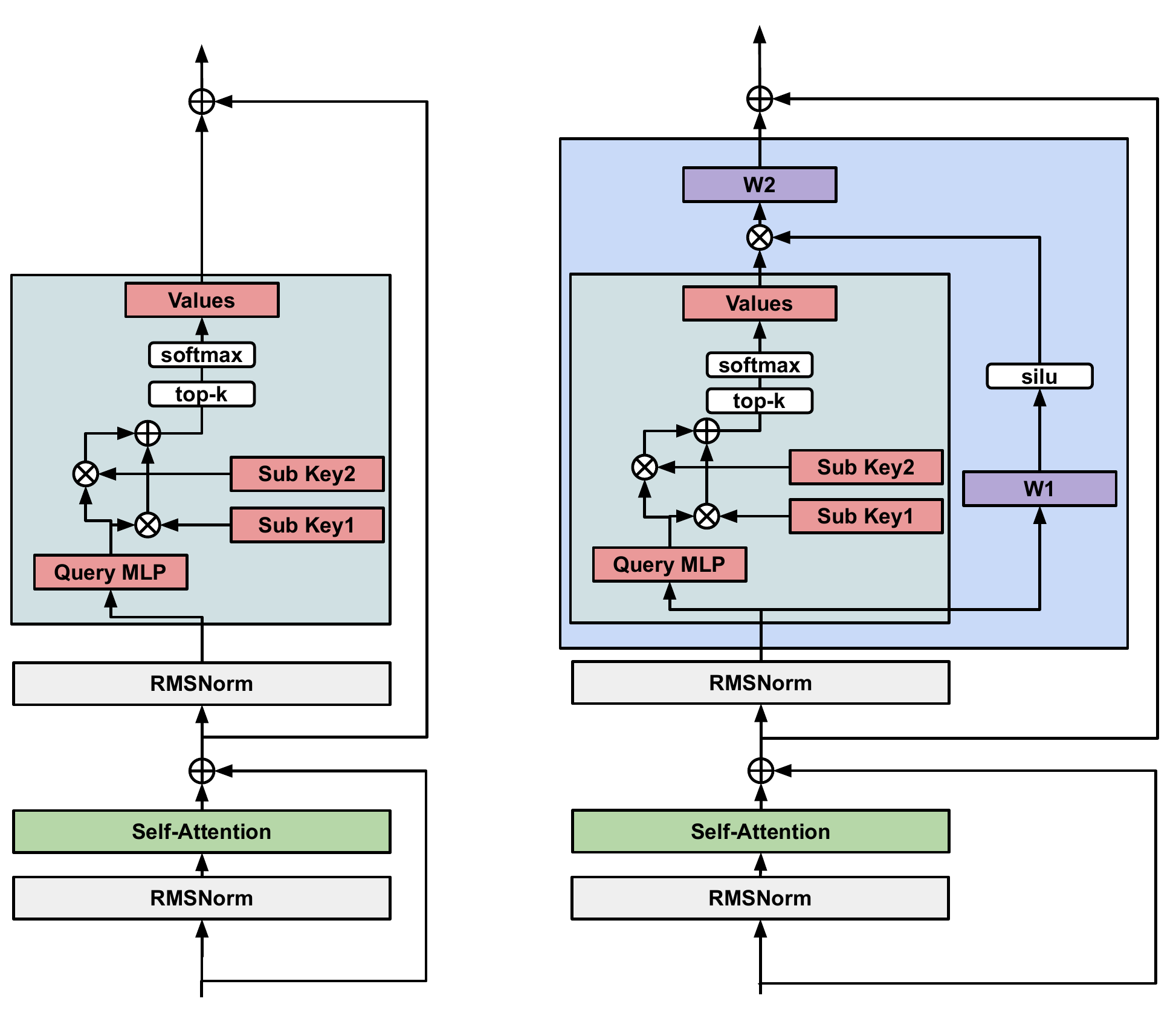}
\caption{On the left the regular memory layer. On the right, the \pkplus{} block, with the added projection, gating and silu non-linearity}
\label{fig:pk_pk_plus}
\end{figure}


We improve training performance of the memory layer by introducing input-dependent gating with a \texttt{silu} non-linearity~\citep{hendrycks2023gaussianerrorlinearunits}.  The output in~\cref{equation1} then becomes
\begin{equation}
    \text{output} =  (y\odot \text{silu}(x^T W_1))^T W_2
\end{equation}
where $\text{silu}(x) = x \ \text{sigmoid} (x)$ and $\odot$ is the element-wise multiplication(see also~\cref{fig:pk_pk_plus}).  We find that for large memory layers, training can become unstable, especially for small base models.  We use qk-normalization~\citep{chameleonteam2024chameleonmixedmodalearlyfusionfoundation} when needed to alleviate this issue.




\section{Experimental setup}\label{experiments}
For our base model architecture, we follow closely the Llama series of dense transformers~\citep{touvron2023llama2openfoundation,dubey2024llama3herdmodels}, which also serve as our dense baselines.  We augment the base models by replacing one or more of the feed-forward layers with a shared memory layer.  For scaling law experiments, we pick base model sizes of 134m, 373m, 720m, and 1.3b parameters.  For these models, we use the Llama2 tokenizer with 32k tokens, and train to 1T tokens with a pretraining data mix that is similar to that of Llama2~\citep{touvron2023llama2openfoundation}.  For experiments at the 8B base model scale, we use the Llama3~\citep{dubey2024llama3herdmodels} configuration and tokenizer (128k tokens), and a better optimized data mix similar to Llama3.

\subsection{Baselines}
In addition to the dense baselines, we also compare to other parameter augmentations including mixture-of-experts (MOE)~\citep{shazeer2017outrageouslylargeneuralnetworks} and the more recent PEER~\citep{he2024mixturemillionexperts} model.  In MOE, each FFN layer is composed of multiple ``experts'', only a subset of which participate in the computation for each input.  The PEER model is conceptually similar to a memory layer, but instead of retrieving a single value embedding, it retrieves a pair of embeddings, which combine into a rank-1 matrix.  Several of these are assembled together into a dynamic feed-forward layer.  PEER works similarly to memory layers in practice, but requires twice the number of parameters for the same number of keys.  Like memory layers, these methods increase the number of parameters in the model without significantly increasing FLOPs.  We pick the number of experts in MOE and the number of keys in PEER to match the number of parameters of our memory-augmented models as closely as possible.  MOE models are trained with expert choice~\citep{zhou2022mixtureofexpertsexpertchoicerouting}, and evaluated with top-1 routing.  PEER layers share the same configuration and hyper-parameters as our memory layer implementation.

\subsection{Evaluation benchmarks}\label{benchmarks}
Our evaluations cover factual question answering (NaturalQuestions~\citep{kwiatkowski-etal-2019-natural}, TriviaQA~\citep{joshi2017triviaqalargescaledistantly}), multi-hop question answering (HotpotQA~\citep{yang2018hotpotqadatasetdiverseexplainable}), scientific and common sense world knowledge (MMLU~\citep{hendrycks2021measuringmassivemultitasklanguage}, HellaSwag~\citep{zellers2019hellaswagmachinereallyfinish}, OBQA~\citep{mihaylov2018suitarmorconductelectricity}, PIQA~\citep{bisk2019piqareasoningphysicalcommonsense}) and coding (HumanEval~\citep{chen2021evaluatinglargelanguagemodels}, MBPP~\citep{austin2021programsynthesislargelanguage}).  We try to report the most commonly used accuracy metrics (exact match or F1 score for QA benchmarks, pass-at-1 for coding).  For some bencmarks, the performance of small models can be very low, and accuracy numbers noisy.  Therefore we use negative log-likelihood (nll) of the correct answer for model ablations.

\section{Scaling results}\label{scaling}
We compare \pk{} models to baselines in a compute-controlled setting.

\subsection{With fixed memory size}
First, we fix the size of the memory, and therefore the number of extra parameters, and compare with the dense baseline, as well as roughly parameter matched MOE and PEER models.  Models with the same base model configuration have negligible differences in FLOPs.  For \pk{} models, we fix the number of half keys to $2^{10}$, and thus the number of memory values to $2^{20}$ (roughly 1 million).  For the PEER baseline, we pick the number of half-keys to be $768$, resulting in slightly more total parameters than \pk{}.  For MOE models, we pick the lowest number of experts such that the parameter count exceeds that of \pk{}.  This corresponds to 16, 8, 6, and 4 experts for the 134m, 373m, 720m and 1.3b sizes respectively.  

The vanilla \pk{} model has a single memory layer, which we pick to replace the middle FFN layer of the transformer.  Our improved \pkplus{} model has 3 memory layers, placed centered with a stride of 4 for the 134m models and 8 for the others.  Additionally it includes a custom swilu non-linearity, and optimized key dimension (set to equal half of the value dim).  As noted earlier, memory layers share parameters, thus have identical memory footprint to a single memory layer.

\begin{figure}[h!]
    \centering
    \begin{minipage}{0.49\textwidth}
        \centering
        \begin{tikzpicture}
            \begin{axis}[
                width=6cm, height=6cm,
                xlabel={Base Parameters},
                ylabel={Accuracy (\%)},
                xmode=log,
                log basis x={10},
                xmin=0, xmax=1400,
                ymin=0, ymax=15,
                legend pos=north west,
                grid=major,
                title={NaturalQuestions},
                xtick={134, 373, 720, 1300},
                xticklabels={134m, 373m, 720m, 1.3b}
            ]
            \addplot[color=blue, mark=square*, line width=1pt] 
            coordinates {
                (134,3.16)
                (373,5.76)
                (720, 9.39)
                (1300,13.68)
            };
            \addlegendentry{Memory+}
            
            \addplot[color=red, mark=triangle*, line width=1pt] 
            coordinates {
                (134,2.49)
                (373,3.99)
                (720,7.04)
                (1300,8.14)
            };
            \addlegendentry{MOE}
            
            \addplot[color=green!60!black, mark=o, line width=1pt] 
            coordinates {
                (134,0.91)
                (373,2.58)
                (720,3.77)
                (1300,7.76)
            };
            \addlegendentry{Dense}
            
            \end{axis}
        \end{tikzpicture}
    \end{minipage}
    \hfill
    \begin{minipage}{0.49\textwidth}
        \centering
        \begin{tikzpicture}
            \begin{axis}[
                width=6cm, height=6cm,
                xlabel={Base Parameters},
                ylabel={f1-score},
                xmode=log,
                log basis x={10},
                xmin=0, xmax=1400,
                ymin=0, ymax=45,
                legend pos=south east,
                grid=major,
                title={TriviaQA},
                xtick={124, 373, 720, 1300},
                xticklabels={124m, 373m, 720m, 1.3b}
            ]
            \addplot[color=blue, mark=square*, line width=1pt] 
            coordinates {
                (134,18.77)
                (373,28.10)
                (720, 36.67)
                (1300,42.89)
            };
            \addlegendentry{Memory+}
            
            \addplot[color=red, mark=triangle*, line width=1pt] 
            coordinates {
                (134,13.08)
                (373,19.99)
                (720,28.07)
                (1300,31.45)
            };
            \addlegendentry{MOE}
            
            \addplot[color=green!60!black, mark=o, line width=1pt] 
            coordinates {
                (134,7.7)
                (373,17.68)
                (720,24.85)
                (1300,32.64)
            };
            \addlegendentry{Dense}
            
            \end{axis}
        \end{tikzpicture}
    \end{minipage}
    \caption{Accuracy vs. Base Parameters for NaturalQuestions and TriviaQA (Memory+ models use 1 million memory embeddings.)}
    \label{fig:qa_figure}
\end{figure}
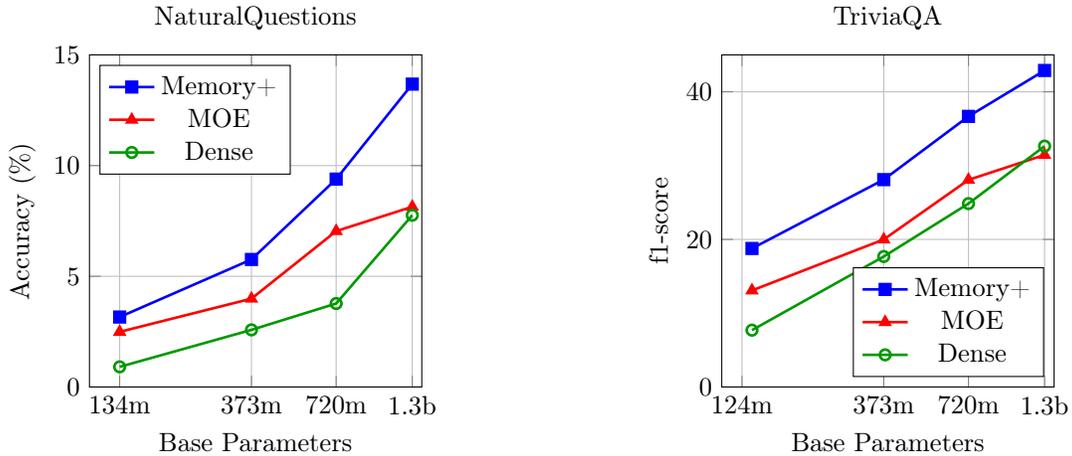

We can see from~\cref{tab:main_results} that \pk{} models improve drastically over the dense baselines, and generally match the performance of models with twice the number of dense parameters on QA tasks.  \pkplus{} improves further over \pk{}, with performance falling generally between dense models with 2x-4x higher compute.  The PEER architecture performs similarly to \pk{} for the same number of parameters, while lagging behind \pkplus{}.  MOE models underperform the memory variants by large margins. \cref{fig:qa_figure} shows the scaling performance of \pk{}, MOE and dense models on QA tasks across various base model sizes.

\subsection{Scaling memory size with a fixed base model}
Next, we investigate scaling behaviour with respect to the memory size for a fixed base model.  In~\cref{fig:memory_scaling}, we see that factual QA performance for a \pkplus{} model keeps increasing predictably with increasing memory size.  At 64 million keys (128 billion memory parameters), a 1.3b \pk{} model approaches the performance of the Llama2 7B model, that has been trained on 2x more tokens using 10x more FLOPs. (see also~\cref{tab:7b_results}).
\begin{table}[h!]
\centering
\caption{Comparing memory augmented architectures with baseline models on QA tasks.  \pk{} models have 1 million value embeddings unless otherwise specified in the model configuration column.  Metrics are accuracy for NQ, PIQA, OBQA and F1 score for TQA, HotpotQA.}
\begin{tabular}{lcllccccc}
\toprule
 & \textbf{Model Configuration} & \textbf{Total Params}  & \textbf{NQ} & \textbf{TQA} & \textbf{PIQA} & \textbf{OBQA} & \textbf{HotPot} \\
\midrule
\multirow{5}{*}{134m} & Dense & 134m & 0.91 & 7.7 & 62.13 & 16.40 & 5.18 \\
                      & MOE   & 984m &  2.49 & 13.08 & 65.78 & 18.80 & 7.80 \\
                      & PEER  & 1.037b& 2.46 & 16.34 & \textbf{67.25} & 17.40 & 8.82 \\
                      & \pk{}    & 937m  &  2.1 & 16.31 & 66.65 & \textbf{17.80} & 9.28 \\
                      & \pkplus{}   & 937m  &  \textbf{3.16} & \textbf{18.77} & 65.94 & 17.60 & \textbf{9.35} \\
\midrule
\multirow{5}{*}{373m} & Dense & 373m &  2.58 & 17.68 & 67.47 & 18.80 & 10.06 \\
                      & MOE   & 1.827b &  3.99 & 19.94 & 68.88 & \textbf{22.20} & 12.50 \\
                      & PEER  & 1.575b &  5.1 & 26.39 & 70.19 & 21.60 & 12.96 \\
                      & \pk{}    & 1.441b &  4.95 & 24.24 & 69.37 & 20.40 & 12.53 \\
                      & \pkplus{}   & 1.434b &  \textbf{5.76} & \textbf{28.10} & \textbf{71.22} & 22.00 & \textbf{13.34} \\
\midrule
\multirow{5}{*}{720m} & Dense & 720m  &  3.77 & 24.85 & 71.33 & 22.60 & 12.90 \\
                      & MOE   & 2.768b  &  7.04 & 28.08 & 70.08 & 20.80 & 14.10 \\
                      & PEER  & 2.517b  &  7.92 & 33.26 & 71.98 & \textbf{25.00} & 14.03 \\
                      & \pk{}    & 2.316b  & 7.2 & 34.8 & 71.82 & 24.40 & \textbf{14.94} \\
                      & \pkplus{}   & 2.316b  &  \textbf{9.39} & \textbf{36.67} & \textbf{72.42} & 24.00 & 14.92 \\
\midrule
\multirow{5}{*}{1.3b} & Dense & 1.3b  &  7.76 & 32.64 & 72.74 & 23.40 & 13.92 \\
                      & MOE   & 3.545b &  8.14 & 31.46 & 73.72 & 25.20 & 15.15 \\
                      & PEER  & 3.646b  &  12.33 & 42.46 & 73.34 & 26.60 & 15.39 \\
                      & \pk{}    & 3.377b &  9.83 & 39.47 & 72.29 & 25.80 & 15.46 \\
                      & \pkplus{}   & 3.377b   &  \textbf{13.68} & \textbf{42.89} & \textbf{75.35} & \textbf{26.80} & \textbf{16.72} \\
\cdashlinelr{2-8}
                      & \pkplus{} 4m   & 9.823b &  14.43 & 51.18 & 75.03 & 27.80 & 18.59 \\
                      & \pkplus{} 16m  & 35.618b   &  20.14 & 58.67 & 76.39 & 26.80 & \textbf{20.65} \\
                      & \pkplus{} 64m  & 138.748b   &  \textbf{20.78} & \textbf{62.14} & \textbf{77.31} & \textbf{30.00} & 20.47 \\
                      
 \midrule
\textit{llama2 7B (2T)} & Dense & 7b & 25.10 & 64.00 & 78.40  & 33.20 & 25.00 \\
\bottomrule
\end{tabular}
\label{tab:main_results}
\end{table}

\subsection{Results at 8B scale}

Finally, we scale our \pkplus{} model with an 8B base model and $4096^2$ memory values (64B memory parameters).  We use the Llama3 8B~\citep{dubey2024llama3herdmodels} architecture and tokenizer, and train on a data mix similar to Llama3~\citep{dubey2024llama3herdmodels}.  We report results at 200 billion and 1 trillion tokens of training in~\cref{tab:7b_results}.  On an expanded set of benchmarks, including general scientific and world knowledge and coding, we see that memory augmented models significantly outperform dense baselines.  The gains are more pronounced earlier in training (200B tokens), suggesting that memory helps models learn facts faster.  At only 1 trillion tokens of training, our \pkplus{} model approaches the performance of Llama3.1 8B, which was trained on 15 trillion tokens.

\begin{table}[h!]
\centering
\caption{Results with an 8B base model.  \pkplus{} models have 16 million memory values (64 billion extra parameters).  Metrics are accuracy for NQ, PIQA, OBQA, HellaSwag, MMLU; F1 score for TQA, HotPotQA; pass@1 for HumanEval, MBPP.  The number of training tokens for each model is denoted in paranthesis.}
\resizebox{\textwidth}{!}{\begin{tabular}{lcccccccccc}
\toprule
 \textbf{Model}  & \textbf{HellaS.} & \textbf{Hotpot} & \textbf{HumanE.} & \textbf{MBPP} & \textbf{MMLU} & \textbf{NQ} & \textbf{OBQA} & \textbf{PIQA} & \textbf{TQA} \\
 \midrule
\textit{llama3.1 8B (15T)} & 60.05 & 27.85 & 37.81 & 48.20 & 66.00 & 29.45 & 34.60 & 79.16 & 70.36 \\
\midrule


dense (200B) & 53.99	& 20.41	& 21.34	& \textbf{30.80}	& 41.35	& 18.61		& \textbf{31.40} 	& 78.02	& 51.741 \\
\pkplus{} (200B) & \textbf{54.33}	& \textbf{21.75} &\textbf{23.17}  	& 29.40	& \textbf{50.14} 	& \textbf{19.36} 	& 30.80	& \textbf{79.11} 	&\textbf{ 57.64} \\
\cdashlinelr{1-10}
dense (1T) & 58.90	& 25.26	& 29.88	& \textbf{44.20}	& 59.68	& 25.24		& 34.20	& \textbf{80.52}	& 63.62 \\
\pkplus{} (1T) & \textbf{60.29}	& \textbf{26.06}	& \textbf{31.71}	& 42.20	& \textbf{63.04}	& \textbf{27.06}		& \textbf{34.40}	& 79.82	& \textbf{68.15} \\

\bottomrule
\end{tabular}}
\label{tab:7b_results}
\end{table}

\subsection{Model ablations}\label{sec:ablations}

In this section, we present results which motivate our modelling choices for the \pkplus{} architecture.
\paragraph{Memory layer placement} 
Since the memory pool is shared, we can replace more FFN layers with memory layers without increasing either the memory or the compute budget.  We see that as we add more memory layers, performance initially increases.  However, as we're effectively removing dense parameters from the model for each added memory layer, eventually the model performance degrades, revealing a sweet spot at around 3 memory layers (\cref{tab:ablation1}, left).  Moreover, we experiment with the placement of these layers, modifying the centring and spacing.  We find that centred placements with larger strides are better, and we adopt this for our \pkplus{} architecture.
\paragraph{Memory layer variants}
We experiment with minor modifications to the memory mechanism (\cref{tab:ablation1}, right).  We try 1. gating the memory with the input using a linear projection, 2. adding a custom swilu non-linearity (\cref{fig:pk_pk_plus}), 3. adding random key-value pairs in addition to the top-k during pre-training to unbias key selection, 4. adding a single fixed key-value pair (softmax sink) to the top-k selected values during pre-training to serve as "anchor".  We find that the swilu non-linearity consistently improves results, and we adopt this improvement into our model.  Simple gating improves performance only in some cases, and swilu already covers this behaviour to some extent, so we decide not to do additional gating.  For key sampling improvements, including the random keys and the fixed (sink) key, we see minor improvements, however these have some negative impact on training speed in our implementation, and the gains were not consistent for larger model sizes, therefore we excluded them from our experiments, leaving this direction open for future exploration.
\begin{table}[ht]
    \centering
    \begin{minipage}{0.49\textwidth}
        \centering
        \begin{tabular}{cccc}
            \toprule
              & \textbf{nll} & \textbf{NQ nll} & \textbf{TQA nll} \\
            \midrule
            \textbf{layer \#} & & & \\
            \midrule
            12  & 2.11 & 12.13 & 8.34 \\
            12,16,20 & 2.08 & 11.60 & 7.54 \\
            8,12,16  & 2.07 & 11.79 & 7.64 \\
            4,12,20  & \textbf{2.06} & \textbf{11.32} & \textbf{7.20} \\
            5,8,11,14,17,21  & 2.11 & 11.79 & 7.73 \\
            \bottomrule
        \end{tabular}
    \end{minipage}%
    \hfill
    \begin{minipage}{0.49\textwidth}
        \centering
        \begin{tabular}{cccc}
            \toprule
             & \textbf{nll} & \textbf{NQ nll} & \textbf{TQA nll} \\
            \midrule
            \textbf{Model} & & & \\
            \midrule
            PK base       & 2.11 & 12.13 & 8.34 \\
            +gated        & 2.11 & 12.24 & 8.17 \\
            +swilu        & 2.11 & 12.05 & 8.09 \\
            +random values & 2.11 & 12.36 & 8.09 \\
            +softmax sink & 2.11 & 12.19 & 8.04 \\
            \bottomrule
        \end{tabular}
    \end{minipage}
    \caption{Ablation studies: on the left, number of memory layers with shared memory, on the right different memory architecture variations.  Metrics are all log likelihood, on the training set, NQ answers and TQA answers. }
    \label{tab:ablation1}
\end{table}

\paragraph{Key and value dimension}
By default, the memory value dimension is chosen to be the same as the base model dimension.  However, we can potentially trade-off the value dimension with the number of values in the memory without changing the total parameter size of the memory using an extra projection after \pk{}.  We present this ablation in~\cref{tab:ablation2}, left, and find that the default configuration is optimal.  We can also independently increase the key embedding dimension, which we do in~\cref{tab:ablation2}, right.  We find unsurprisingly that increasing the key dim is beneficial.  However, increasing the key dim does add more dense parameters to the model, and thus we cannot increase it indefinitely without breaking our fair comparisons.  We pick a key dimension of half the base model dim for our experiments.
\begin{table}[ht]
    \centering
    \begin{minipage}{0.49\textwidth}
        \centering
        \begin{tabular}{ccccc}
            \toprule
             & & \textbf{nll} & \textbf{NQ nll} & \textbf{TQA nll} \\
            \midrule
            \textbf{v\_dim} & \textbf{\#values} & & & \\
            \midrule
            64 & 16m & 2.15 & 12.86 & 8.75 \\
            256 & 4m & 2.14 & 12.63 & 8.49 \\
            1024 & 1m & \textbf{2.11} & \textbf{12.13} & \textbf{8.34} \\
            2048 & 512k & 2.14 & 12.49 & 8.53 \\
            \bottomrule
        \end{tabular}
    \end{minipage}%
    \hfill
    \begin{minipage}{0.49\textwidth}
        \centering
        \begin{tabular}{cccc}
            \toprule
             & \textbf{nll} & \textbf{NQ nll} & \textbf{TQA nll} \\
            \midrule
            \textbf{key\_dim} & & & \\
            \midrule
            256  & 2.11 & 12.13 & 8.34 \\
            512 & 2.12 & 12.32 & 8.15 \\
            1024  & 2.11 & 12.37 & 8.25 \\
            2048  & \textbf{2.09} & \textbf{11.98} & \textbf{7.83} \\
            \bottomrule
        \end{tabular}
    \end{minipage}
    \caption{Ablation studies: on the left, varying the value embedding dim while keeping total parameter count the same, on the right varying key dim.  Metrics are all log likelihood, on the training set, NQ answers and TQA answers. These were ran on the 373m model size, which uses a latent dimension of 1024. key\_dim is the sum of the dimension of the sub-keys.}
    \label{tab:ablation2}
\end{table}

\section{Implications and shortcomings of the work}
Scaling of dense transformer models has dominated progress in the AI field in the last 6 years.  As this scaling is nearing its physical and resource limits, it's useful to consider alternatives which might be equally scalable without being as compute and energy intensive.  Memory layers with their sparse activations nicely complement dense networks, providing increased capacity for knowledge acquisition while being light on compute.  They can be efficiently scaled, and provide practitioners with an attractive new direction to trade-off memory with compute.  

While the memory layer implementation presented here is orders of magnitude more scalable than previous works, there still remains a substantial engineering task to make them efficient enough for large scale production uses.  Dense architectures have been optimized for and co-evolved with modern GPU architectures for decades.  While we believe it's in principle possible to make memory layers as fast, or even faster than regular FFN layers on current hardware, we acknowledge that this needs non-trivial effort.

We have so far presented only high level empirical evidence that memory layers improve factuality of models.  However, we believe the sparse updates made possible by memory layers might have deep implications to how models learn and store information.  In particular, we hope that new learning methods can be developed to push the effectiveness of these layers even further, enabling less forgetting, fewer hallucinations, and continual learning.

\subsubsection*{Acknowledgments}
We would like to thank Francisco Massa, Luca Wehrstedt for valuable input on making memory layers more efficient; Gabriel Synnaeve, Ammer Rizvi, Michel Meyer for helping to provide resources for scaling experiments.

\clearpage
\newpage
\bibliographystyle{assets/plainnat}
\bibliography{paper}

\begin{thebibliography}{44}
\providecommand{\natexlab}[1]{#1}
\providecommand{\url}[1]{\texttt{#1}}
\expandafter\ifx\csname urlstyle\endcsname\relax
  \providecommand{\doi}[1]{doi: #1}\else
  \providecommand{\doi}{doi: \begingroup \urlstyle{rm}\Url}\fi

\bibitem[Austin et~al.(2021)Austin, Odena, Nye, Bosma, Michalewski, Dohan,
  Jiang, Cai, Terry, Le, and Sutton]{austin2021programsynthesislargelanguage}
Jacob Austin, Augustus Odena, Maxwell Nye, Maarten Bosma, Henryk Michalewski,
  David Dohan, Ellen Jiang, Carrie Cai, Michael Terry, Quoc Le, and Charles
  Sutton.
\newblock Program synthesis with large language models, 2021.
\newblock \url{https://arxiv.org/abs/2108.07732}.

\bibitem[Bahdanau et~al.(2016)Bahdanau, Cho, and
  Bengio]{bahdanau2016neuralmachinetranslationjointly}
Dzmitry Bahdanau, Kyunghyun Cho, and Yoshua Bengio.
\newblock Neural machine translation by jointly learning to align and
  translate, 2016.
\newblock \url{https://arxiv.org/abs/1409.0473}.

\bibitem[Bisk et~al.(2019)Bisk, Zellers, Bras, Gao, and
  Choi]{bisk2019piqareasoningphysicalcommonsense}
Yonatan Bisk, Rowan Zellers, Ronan~Le Bras, Jianfeng Gao, and Yejin Choi.
\newblock Piqa: Reasoning about physical commonsense in natural language, 2019.
\newblock \url{https://arxiv.org/abs/1911.11641}.

\bibitem[Brown et~al.(2020)Brown, Mann, Ryder, Subbiah, Kaplan, Dhariwal,
  Neelakantan, Shyam, Sastry, and et.
  al.]{brown2020languagemodelsfewshotlearners}
Tom~B. Brown, Benjamin Mann, Nick Ryder, Melanie Subbiah, Jared Kaplan,
  Prafulla Dhariwal, Arvind Neelakantan, Pranav Shyam, Girish Sastry, and
  Amanda~Askell et. al.
\newblock Language models are few-shot learners, 2020.
\newblock \url{https://arxiv.org/abs/2005.14165}.

\bibitem[Chen and Yih(2020)]{chen-yih-2020-open}
Danqi Chen and Wen-tau Yih.
\newblock Open-domain question answering.
\newblock In Agata Savary and Yue Zhang, editors, \emph{Proceedings of the 58th
  Annual Meeting of the Association for Computational Linguistics: Tutorial
  Abstracts}, pages 34--37, Online, July 2020. Association for Computational
  Linguistics.
\newblock \doi{10.18653/v1/2020.acl-tutorials.8}.
\newblock \url{https://aclanthology.org/2020.acl-tutorials.8}.

\bibitem[Chen et~al.(2017)Chen, Fisch, Weston, and
  Bordes]{chen2017readingwikipediaansweropendomain}
Danqi Chen, Adam Fisch, Jason Weston, and Antoine Bordes.
\newblock Reading wikipedia to answer open-domain questions, 2017.
\newblock \url{https://arxiv.org/abs/1704.00051}.

\bibitem[Chen et~al.(2021)Chen, Tworek, Jun, Yuan, de~Oliveira~Pinto, Kaplan,
  Edwards, Burda, Joseph, Brockman, Ray, Puri, Krueger, Petrov, and et.
  al.]{chen2021evaluatinglargelanguagemodels}
Mark Chen, Jerry Tworek, Heewoo Jun, Qiming Yuan, Henrique~Ponde
  de~Oliveira~Pinto, Jared Kaplan, Harri Edwards, Yuri Burda, Nicholas Joseph,
  Greg Brockman, Alex Ray, Raul Puri, Gretchen Krueger, Michael Petrov, and
  Heidy~Khlaaf et. al.
\newblock Evaluating large language models trained on code, 2021.
\newblock \url{https://arxiv.org/abs/2107.03374}.

\bibitem[Dubey et~al.(2024)Dubey, Jauhri, Pandey, Kadian, Al-Dahle, Letman,
  Mathur, Schelten, Yang, Fan, Goyal, Hartshorn, Yang, Mitra, and et.
  al.]{dubey2024llama3herdmodels}
Abhimanyu Dubey, Abhinav Jauhri, Abhinav Pandey, Abhishek Kadian, Ahmad
  Al-Dahle, Aiesha Letman, Akhil Mathur, Alan Schelten, Amy Yang, Angela Fan,
  Anirudh Goyal, Anthony Hartshorn, Aobo Yang, Archi Mitra, and
  Archie~Sravankumar et. al.
\newblock The llama 3 herd of models, 2024.
\newblock \url{https://arxiv.org/abs/2407.21783}.

\bibitem[Ganguli et~al.(2022)Ganguli, Hernandez, Lovitt, Askell, Bai, Chen,
  Conerly, Dassarma, Drain, Elhage, and El~Showk]{Ganguli_2022}
Deep Ganguli, Danny Hernandez, Liane Lovitt, Amanda Askell, Yuntao Bai, Anna
  Chen, Tom Conerly, Nova Dassarma, Dawn Drain, Nelson Elhage, and Sheer
  et.~al. El~Showk.
\newblock Predictability and surprise in large generative models.
\newblock In \emph{2022 ACM Conference on Fairness, Accountability, and
  Transparency}, FAccT ’22. ACM, June 2022.
\newblock \doi{10.1145/3531146.3533229}.
\newblock \url{http://dx.doi.org/10.1145/3531146.3533229}.

\bibitem[Graves et~al.(2014)Graves, Wayne, and
  Danihelka]{graves2014neuralturingmachines}
Alex Graves, Greg Wayne, and Ivo Danihelka.
\newblock Neural turing machines, 2014.
\newblock \url{https://arxiv.org/abs/1410.5401}.

\bibitem[Graves et~al.(2016)Graves, Wayne, Reynolds, Harley, Danihelka,
  Grabska-Barwińska, Colmenarejo, Grefenstette, Ramalho, Agapiou, Badia,
  Hermann, Zwols, Ostrovski, Cain, King, Summerfield, Blunsom, Kavukcuoglu, and
  Hassabis]{graves2016hybrid}
Alex Graves, Greg Wayne, Malcolm Reynolds, Tim Harley, Ivo Danihelka, Agnieszka
  Grabska-Barwińska, Sergio~Gómez Colmenarejo, Edward Grefenstette, Tiago
  Ramalho, John Agapiou, Adrià~Puigdomènech Badia, Karl~Moritz Hermann, Yori
  Zwols, Georg Ostrovski, Adam Cain, Helen King, Christopher Summerfield, Phil
  Blunsom, Koray Kavukcuoglu, and Demis Hassabis.
\newblock Hybrid computing using a neural network with dynamic external memory.
\newblock \emph{Nature}, 538\penalty0 (7626):\penalty0 471--476, October 2016.
\newblock ISSN 00280836.
\newblock \url{http://dx.doi.org/10.1038/nature20101}.

\bibitem[Guu et~al.(2020)Guu, Lee, Tung, Pasupat, and
  Chang]{guu2020realmretrievalaugmentedlanguagemodel}
Kelvin Guu, Kenton Lee, Zora Tung, Panupong Pasupat, and Ming-Wei Chang.
\newblock Realm: Retrieval-augmented language model pre-training, 2020.
\newblock \url{https://arxiv.org/abs/2002.08909}.

\bibitem[He(2024)]{he2024mixturemillionexperts}
Xu~Owen He.
\newblock Mixture of a million experts, 2024.
\newblock \url{https://arxiv.org/abs/2407.04153}.

\bibitem[Hendrycks and Gimpel(2023)]{hendrycks2023gaussianerrorlinearunits}
Dan Hendrycks and Kevin Gimpel.
\newblock Gaussian error linear units (gelus), 2023.
\newblock \url{https://arxiv.org/abs/1606.08415}.

\bibitem[Hendrycks et~al.(2021)Hendrycks, Burns, Basart, Zou, Mazeika, Song,
  and Steinhardt]{hendrycks2021measuringmassivemultitasklanguage}
Dan Hendrycks, Collin Burns, Steven Basart, Andy Zou, Mantas Mazeika, Dawn
  Song, and Jacob Steinhardt.
\newblock Measuring massive multitask language understanding, 2021.
\newblock \url{https://arxiv.org/abs/2009.03300}.

\bibitem[Hornik et~al.(1989)Hornik, Stinchcombe, and White]{HORNIK1989359}
Kurt Hornik, Maxwell Stinchcombe, and Halbert White.
\newblock Multilayer feedforward networks are universal approximators.
\newblock \emph{Neural Networks}, 2\penalty0 (5):\penalty0 359--366, 1989.
\newblock ISSN 0893-6080.
\newblock \doi{https://doi.org/10.1016/0893-6080(89)90020-8}.
\newblock
  \url{https://www.sciencedirect.com/science/article/pii/0893608089900208}.

\bibitem[Ji et~al.(2023)Ji, Lee, Frieske, Yu, Su, Xu, Ishii, Bang, Madotto, and
  Fung]{Ji_2023}
Ziwei Ji, Nayeon Lee, Rita Frieske, Tiezheng Yu, Dan Su, Yan Xu, Etsuko Ishii,
  Ye~Jin Bang, Andrea Madotto, and Pascale Fung.
\newblock Survey of hallucination in natural language generation.
\newblock \emph{ACM Computing Surveys}, 55\penalty0 (12):\penalty0 1–38,
  March 2023.
\newblock ISSN 1557-7341.
\newblock \doi{10.1145/3571730}.
\newblock \url{http://dx.doi.org/10.1145/3571730}.

\bibitem[Jiang et~al.(2024)Jiang, Sablayrolles, Roux, Mensch, Savary, Bamford,
  Chaplot, de~las Casas, Hanna, Bressand, Lengyel, Bour, and et.
  al.]{jiang2024mixtralexperts}
Albert~Q. Jiang, Alexandre Sablayrolles, Antoine Roux, Arthur Mensch, Blanche
  Savary, Chris Bamford, Devendra~Singh Chaplot, Diego de~las Casas, Emma~Bou
  Hanna, Florian Bressand, Gianna Lengyel, Guillaume Bour, and Guillaume~Lample
  et. al.
\newblock Mixtral of experts, 2024.
\newblock \url{https://arxiv.org/abs/2401.04088}.

\bibitem[Johnson et~al.(2019)Johnson, Douze, and J{\'e}gou]{johnson2019billion}
Jeff Johnson, Matthijs Douze, and Herv{\'e} J{\'e}gou.
\newblock Billion-scale similarity search with {GPUs}.
\newblock \emph{IEEE Transactions on Big Data}, 7\penalty0 (3):\penalty0
  535--547, 2019.

\bibitem[Joshi et~al.(2017)Joshi, Choi, Weld, and
  Zettlemoyer]{joshi2017triviaqalargescaledistantly}
Mandar Joshi, Eunsol Choi, Daniel~S. Weld, and Luke Zettlemoyer.
\newblock Triviaqa: A large scale distantly supervised challenge dataset for
  reading comprehension, 2017.
\newblock \url{https://arxiv.org/abs/1705.03551}.

\bibitem[Kaplan et~al.(2020)Kaplan, McCandlish, Henighan, Brown, Chess, Child,
  Gray, Radford, Wu, and Amodei]{kaplan2020scalinglawsneurallanguage}
Jared Kaplan, Sam McCandlish, Tom Henighan, Tom~B. Brown, Benjamin Chess, Rewon
  Child, Scott Gray, Alec Radford, Jeffrey Wu, and Dario Amodei.
\newblock Scaling laws for neural language models, 2020.
\newblock \url{https://arxiv.org/abs/2001.08361}.

\bibitem[Karpukhin et~al.(2020)Karpukhin, Oğuz, Min, Lewis, Wu, Edunov, Chen,
  and tau Yih]{karpukhin2020densepassageretrievalopendomain}
Vladimir Karpukhin, Barlas Oğuz, Sewon Min, Patrick Lewis, Ledell Wu, Sergey
  Edunov, Danqi Chen, and Wen tau Yih.
\newblock Dense passage retrieval for open-domain question answering, 2020.
\newblock \url{https://arxiv.org/abs/2004.04906}.

\bibitem[Khandelwal et~al.(2020)Khandelwal, Levy, Jurafsky, Zettlemoyer, and
  Lewis]{khandelwal2020generalizationmemorizationnearestneighbor}
Urvashi Khandelwal, Omer Levy, Dan Jurafsky, Luke Zettlemoyer, and Mike Lewis.
\newblock Generalization through memorization: Nearest neighbor language
  models, 2020.
\newblock \url{https://arxiv.org/abs/1911.00172}.

\bibitem[Kwiatkowski et~al.(2019)Kwiatkowski, Palomaki, Redfield, Collins,
  Parikh, Alberti, Epstein, Polosukhin, Devlin, Lee, Toutanova, Jones, Kelcey,
  Chang, Dai, Uszkoreit, Le, and Petrov]{kwiatkowski-etal-2019-natural}
Tom Kwiatkowski, Jennimaria Palomaki, Olivia Redfield, Michael Collins, Ankur
  Parikh, Chris Alberti, Danielle Epstein, Illia Polosukhin, Jacob Devlin,
  Kenton Lee, Kristina Toutanova, Llion Jones, Matthew Kelcey, Ming-Wei Chang,
  Andrew~M. Dai, Jakob Uszkoreit, Quoc Le, and Slav Petrov.
\newblock Natural questions: A benchmark for question answering research.
\newblock \emph{Transactions of the Association for Computational Linguistics},
  7:\penalty0 452--466, 2019.
\newblock \doi{10.1162/tacl_a_00276}.
\newblock \url{https://aclanthology.org/Q19-1026}.

\bibitem[Lample et~al.(2019)Lample, Sablayrolles, Ranzato, Denoyer, and
  Jégou]{lample2019largememorylayersproduct}
Guillaume Lample, Alexandre Sablayrolles, Marc'Aurelio Ranzato, Ludovic
  Denoyer, and Hervé Jégou.
\newblock Large memory layers with product keys, 2019.
\newblock \url{https://arxiv.org/abs/1907.05242}.

\bibitem[Lee et~al.(2019)Lee, Chang, and
  Toutanova]{lee2019latentretrievalweaklysupervised}
Kenton Lee, Ming-Wei Chang, and Kristina Toutanova.
\newblock Latent retrieval for weakly supervised open domain question
  answering, 2019.
\newblock \url{https://arxiv.org/abs/1906.00300}.

\bibitem[Lepikhin et~al.(2020)Lepikhin, Lee, Xu, Chen, Firat, Huang, Krikun,
  Shazeer, and Chen]{lepikhin2020gshardscalinggiantmodels}
Dmitry Lepikhin, HyoukJoong Lee, Yuanzhong Xu, Dehao Chen, Orhan Firat, Yanping
  Huang, Maxim Krikun, Noam Shazeer, and Zhifeng Chen.
\newblock Gshard: Scaling giant models with conditional computation and
  automatic sharding, 2020.
\newblock \url{https://arxiv.org/abs/2006.16668}.

\bibitem[Lewis et~al.(2021)Lewis, Perez, Piktus, Petroni, Karpukhin, Goyal,
  Küttler, Lewis, tau Yih, Rocktäschel, Riedel, and
  Kiela]{lewis2021retrievalaugmentedgenerationknowledgeintensivenlp}
Patrick Lewis, Ethan Perez, Aleksandra Piktus, Fabio Petroni, Vladimir
  Karpukhin, Naman Goyal, Heinrich Küttler, Mike Lewis, Wen tau Yih, Tim
  Rocktäschel, Sebastian Riedel, and Douwe Kiela.
\newblock Retrieval-augmented generation for knowledge-intensive nlp tasks,
  2021.
\newblock \url{https://arxiv.org/abs/2005.11401}.

\bibitem[Mihaylov et~al.(2018)Mihaylov, Clark, Khot, and
  Sabharwal]{mihaylov2018suitarmorconductelectricity}
Todor Mihaylov, Peter Clark, Tushar Khot, and Ashish Sabharwal.
\newblock Can a suit of armor conduct electricity? a new dataset for open book
  question answering, 2018.
\newblock \url{https://arxiv.org/abs/1809.02789}.

\bibitem[OpenAI et~al.(2024)OpenAI, Achiam, Adler, Agarwal, Ahmad, Akkaya,
  Aleman, Almeida, Altenschmidt, Altman, Anadkat, Avila, Babuschkin, and et.
  al.]{openai2024gpt4technicalreport}
OpenAI, Josh Achiam, Steven Adler, Sandhini Agarwal, Lama Ahmad, Ilge Akkaya,
  Florencia~Leoni Aleman, Diogo Almeida, Janko Altenschmidt, Sam Altman,
  Shyamal Anadkat, Red Avila, Igor Babuschkin, and Suchir~Balaji et. al.
\newblock Gpt-4 technical report, 2024.
\newblock \url{https://arxiv.org/abs/2303.08774}.

\bibitem[Petroni et~al.(2021)Petroni, Piktus, Fan, Lewis, Yazdani, Cao, Thorne,
  Jernite, Karpukhin, Maillard, Plachouras, Rocktäschel, and
  Riedel]{petroni2021kiltbenchmarkknowledgeintensive}
Fabio Petroni, Aleksandra Piktus, Angela Fan, Patrick Lewis, Majid Yazdani,
  Nicola~De Cao, James Thorne, Yacine Jernite, Vladimir Karpukhin, Jean
  Maillard, Vassilis Plachouras, Tim Rocktäschel, and Sebastian Riedel.
\newblock Kilt: a benchmark for knowledge intensive language tasks, 2021.
\newblock \url{https://arxiv.org/abs/2009.02252}.

\bibitem[Roberts et~al.(2020)Roberts, Raffel, and
  Shazeer]{roberts2020knowledgepackparameterslanguage}
Adam Roberts, Colin Raffel, and Noam Shazeer.
\newblock How much knowledge can you pack into the parameters of a language
  model?, 2020.
\newblock \url{https://arxiv.org/abs/2002.08910}.

\bibitem[Schaeffer et~al.(2023)Schaeffer, Miranda, and
  Koyejo]{schaeffer2023emergentabilitieslargelanguage}
Rylan Schaeffer, Brando Miranda, and Sanmi Koyejo.
\newblock Are emergent abilities of large language models a mirage?, 2023.
\newblock \url{https://arxiv.org/abs/2304.15004}.

\bibitem[Shazeer et~al.(2017)Shazeer, Mirhoseini, Maziarz, Davis, Le, Hinton,
  and Dean]{shazeer2017outrageouslylargeneuralnetworks}
Noam Shazeer, Azalia Mirhoseini, Krzysztof Maziarz, Andy Davis, Quoc Le,
  Geoffrey Hinton, and Jeff Dean.
\newblock Outrageously large neural networks: The sparsely-gated
  mixture-of-experts layer, 2017.
\newblock \url{https://arxiv.org/abs/1701.06538}.

\bibitem[Sukhbaatar et~al.(2015)Sukhbaatar, szlam, Weston, and
  Fergus]{sukhbaatar2015endtoendmemorynetworks}
Sainbayar Sukhbaatar, arthur szlam, Jason Weston, and Rob Fergus.
\newblock End-to-end memory networks.
\newblock In C.~Cortes, N.~Lawrence, D.~Lee, M.~Sugiyama, and R.~Garnett,
  editors, \emph{Advances in Neural Information Processing Systems}, volume~28.
  Curran Associates, Inc., 2015.
\newblock
  \url{https://proceedings.neurips.cc/paper_files/paper/2015/file/8fb21ee7a2207526da55a679f0332de2-Paper.pdf}.

\bibitem[Team(2024)]{chameleonteam2024chameleonmixedmodalearlyfusionfoundation}
Chameleon Team.
\newblock Chameleon: Mixed-modal early-fusion foundation models, 2024.
\newblock \url{https://arxiv.org/abs/2405.09818}.

\bibitem[Team et~al.(2024)Team, Georgiev, Lei, Burnell, Bai, Gulati, Tanzer,
  Vincent, Pan, Wang, Mariooryad, Ding, Geng, Alcober, and et.
  al.]{geminiteam2024gemini15unlockingmultimodal}
Gemini Team, Petko Georgiev, Ving~Ian Lei, Ryan Burnell, Libin Bai, Anmol
  Gulati, Garrett Tanzer, Damien Vincent, Zhufeng Pan, Shibo Wang, Soroosh
  Mariooryad, Yifan Ding, Xinyang Geng, Fred Alcober, and Roy~Frostig et. al.
\newblock Gemini 1.5: Unlocking multimodal understanding across millions of
  tokens of context, 2024.
\newblock \url{https://arxiv.org/abs/2403.05530}.

\bibitem[Touvron et~al.(2023)Touvron, Martin, Stone, Albert, Almahairi, Babaei,
  Bashlykov, Batra, Bhargava, Bhosale, Bikel, Blecher, Ferrer, Chen, and et.
  al.]{touvron2023llama2openfoundation}
Hugo Touvron, Louis Martin, Kevin Stone, Peter Albert, Amjad Almahairi, Yasmine
  Babaei, Nikolay Bashlykov, Soumya Batra, Prajjwal Bhargava, Shruti Bhosale,
  Dan Bikel, Lukas Blecher, Cristian~Canton Ferrer, Moya Chen, and
  Guillem~Cucurull et. al.
\newblock Llama 2: Open foundation and fine-tuned chat models, 2023.
\newblock \url{https://arxiv.org/abs/2307.09288}.

\bibitem[Vaswani et~al.(2023)Vaswani, Shazeer, Parmar, Uszkoreit, Jones, Gomez,
  Kaiser, and Polosukhin]{vaswani2023attentionneed}
Ashish Vaswani, Noam Shazeer, Niki Parmar, Jakob Uszkoreit, Llion Jones,
  Aidan~N. Gomez, Lukasz Kaiser, and Illia Polosukhin.
\newblock Attention is all you need, 2023.
\newblock \url{https://arxiv.org/abs/1706.03762}.

\bibitem[Wei et~al.(2022)Wei, Tay, Bommasani, Raffel, Zoph, Borgeaud, Yogatama,
  Bosma, Zhou, Metzler, Chi, Hashimoto, Vinyals, Liang, Dean, and
  Fedus]{wei2022emergentabilitieslargelanguage}
Jason Wei, Yi~Tay, Rishi Bommasani, Colin Raffel, Barret Zoph, Sebastian
  Borgeaud, Dani Yogatama, Maarten Bosma, Denny Zhou, Donald Metzler, Ed~H.
  Chi, Tatsunori Hashimoto, Oriol Vinyals, Percy Liang, Jeff Dean, and William
  Fedus.
\newblock Emergent abilities of large language models, 2022.
\newblock \url{https://arxiv.org/abs/2206.07682}.

\bibitem[Weston et~al.(2015)Weston, Chopra, and
  Bordes]{weston2015memorynetworks}
Jason Weston, Sumit Chopra, and Antoine Bordes.
\newblock Memory networks, 2015.
\newblock \url{https://arxiv.org/abs/1410.3916}.

\bibitem[Yang et~al.(2018)Yang, Qi, Zhang, Bengio, Cohen, Salakhutdinov, and
  Manning]{yang2018hotpotqadatasetdiverseexplainable}
Zhilin Yang, Peng Qi, Saizheng Zhang, Yoshua Bengio, William~W. Cohen, Ruslan
  Salakhutdinov, and Christopher~D. Manning.
\newblock Hotpotqa: A dataset for diverse, explainable multi-hop question
  answering, 2018.
\newblock \url{https://arxiv.org/abs/1809.09600}.

\bibitem[Zellers et~al.(2019)Zellers, Holtzman, Bisk, Farhadi, and
  Choi]{zellers2019hellaswagmachinereallyfinish}
Rowan Zellers, Ari Holtzman, Yonatan Bisk, Ali Farhadi, and Yejin Choi.
\newblock Hellaswag: Can a machine really finish your sentence?, 2019.
\newblock \url{https://arxiv.org/abs/1905.07830}.

\bibitem[Zhou et~al.(2022)Zhou, Lei, Liu, Du, Huang, Zhao, Dai, Chen, Le, and
  Laudon]{zhou2022mixtureofexpertsexpertchoicerouting}
Yanqi Zhou, Tao Lei, Hanxiao Liu, Nan Du, Yanping Huang, Vincent Zhao, Andrew
  Dai, Zhifeng Chen, Quoc Le, and James Laudon.
\newblock Mixture-of-experts with expert choice routing, 2022.
\newblock \url{https://arxiv.org/abs/2202.09368}.

\end{thebibliography}



\end{document}